\documentclass[a4paper, 
10pt
]{article}

\usepackage[a4paper,left=3cm,right=3cm,top=2.5cm,bottom=2.5cm]{geometry}
\usepackage{amsmath}
\usepackage{multirow}
\usepackage[table]{xcolor}
\usepackage{amsopn}
\usepackage{mathtools}
\usepackage{comment}
\usepackage{booktabs}
\usepackage{algorithm}
\usepackage{algpseudocode}
\usepackage{amssymb}
\usepackage{url}
\usepackage{bm}
\usepackage{tabularx}
\usepackage{enumerate}
\usepackage{array}
\usepackage{tikz}
\usepackage[export]{adjustbox}
\usepackage{tikzsymbols}
\usepackage{authblk}
\numberwithin{equation}{subsection}
\usepackage{xcolor}
\usepackage[hidelinks]{hyperref}
\usepackage{cleveref}
\usepackage{chngcntr}

\counterwithin*{equation}{section}
\counterwithin*{equation}{subsection}
\counterwithin*{equation}{subsubsection}
\counterwithin*{table}{section}
\counterwithin*{table}{subsection}
\counterwithin*{table}{subsubsection}

\newcommand{\reviewerA}[1]{{\color{black}#1}}
\newcommand{\reviewerB}[1]{{\color{black}#1}}

\begin{document}

\title{A Continuous Convolutional Trainable Filter for Modelling Unstructured Data}
\author[1]{Dario Coscia\footnote{dario.coscia@sissa.it}}
\author[1]{Laura~Meneghetti\footnote{laura.meneghetti@sissa.it}}
\author[1]{Nicola~Demo\footnote{nicola.demo@sissa.it}}
\author[2,1]{Giovanni~Stabile\footnote{giovanni.stabile@uniurb.it}}
\author[1]{Gianluigi~Rozza\footnote{gianluigi.rozza@sissa.it}}

\affil[1]{Mathematics Area, mathLab, SISSA, via Bonomea 265, I-34136, Trieste, Italy}
\affil[2]{Department of Pure and Applied Sciences, Informatics and Mathematics Section, University of Urbino Carlo Bo, Piazza della Repubblica 13, I-61029, Urbino, Italy}
\maketitle

\begin{abstract}
   Convolutional Neural Network (CNN) is one of the most important architectures in deep learning. The fundamental building block of a CNN is a trainable filter, represented as a discrete grid, used to perform convolution on discrete input data. In this work, we propose a continuous version of a trainable convolutional filter able to work also with unstructured data. This new framework allows exploring CNNs beyond discrete domains, enlarging the usage of this important learning technique for many more complex problems. Our experiments show that the continuous filter can achieve a level of accuracy comparable to the state-of-the-art discrete filter, and that it can be used in current deep learning architectures as a building block to solve problems with unstructured domains as well. 
\end{abstract}


\section{Introduction}\label{sec1}
In the deep learning field, a convolutional neural network (CNN)~\cite{lecun1989generalization} is one of the most important architectures, widely used in academia and industrial research. For an overview of the topic, the interested reader might refer to \cite{CNNreview, Goodfellow-et-al-2016, alzubaidi2021review, calin2020deep, zhang2021dive}. Despite the great success in many fields including, but not limited, to computer vision~\cite{krizhevsky2012imagenet, shanmugamani2018deep, jiang2019deep} or natural language processing~\cite{young2018recent, deng2018deep}, current CNNs are constrained to structural data. Indeed, the basic building block of a CNN is a trainable filter, represented by a discrete grid, which performs cross-correlation, also known as convolution, on a discrete domain. Nevertheless, the idea behind convolution can be easily extended mathematically to unstructured domains, for reference see \cite{Heil2019_Convolution_maths}. 
\reviewerA{One possible approach for this kind of problem is the graph neural networks (GNN)~\cite{kipf2017semisupervised,9046288}, where a graph is built starting from the topology of the discretized space. This allows us to apply convolution even to unstructured data by looking at the graph edges, bypassing in this way the limitations of the standard CNNs approach. However, GNNs typically require huge computational resources, due to their implicit complexity.

Instead}
in this article, we present a methodology to apply CNNs to unstructured data by introducing a continuous extension of a convolutional filter, named \textit{continuous filter}, \reviewerA{without modeling the data using a graph}. 
\begin{figure}[th]
    \centering
    \includegraphics[width=0.95\linewidth]{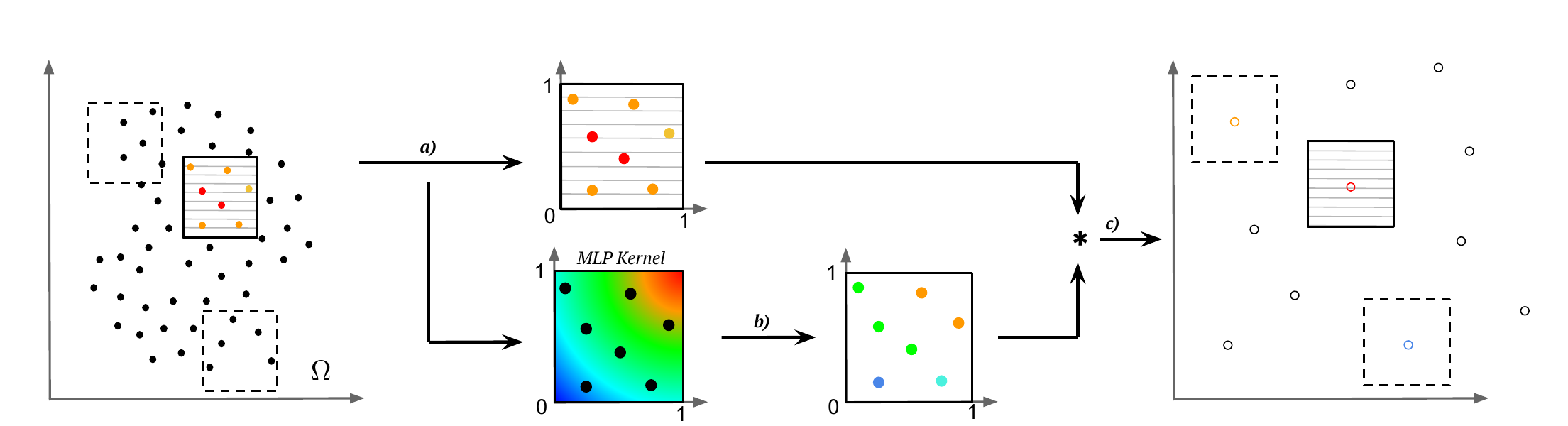}
    \caption{Continuous convolutional filter process. The unstructured domain input points falling into the filter are mapped in the filter domain (a). The filter values are approximated with a MLP kernel (b). Finally, the convolution between the mapped values and the filter values is performed (c).}
    \label{fig:abstract-grafico}
\end{figure}
The main idea, which is depicted graphically in Figure \ref{fig:abstract-grafico}, relies on approximating the continuous filter with a trainable function using a feed-forward neural network and perform standard continuous convolution between the input data and the continuous filter. Previous works have introduced different approaches to continuous convolution in various settings ranging from informatics and graph neural networks to physics and modeling quantum interactions, see for example \cite{schnet, shocher2020discrete, graph_convolutional_NN}. Even so, the latter is difficult to generalize, and an \reviewerB{analogy} with a discrete CNN filter is not straightforward. To our extent \cite{wang2018deep, romero_et_al} are the closest works in literature to our approach, both approximating the trainable filter function with a feed-forward neural network and performing continuous convolution. However, \cite{wang2018deep} and \cite{romero_et_al} focus on filters with unbounded domains for convolution. In our work, we instead fix the dimension of the filter, as in state of the art discrete filters, and learn the approximation function on the filter domain. This introduces a neat \reviewerB{analogy} to discrete CNN filters. Furthermore, differently from \cite{wang2018deep, romero_et_al}, we also cover important properties of convolution, such as transposed convolution or different approaches to multichannel convolution. To summarize, in this work we aim to reproduce as closely as possible a discrete CNN filter but in a continuous not structured domain setting, in order to exploit the main deep learning architectures, based on CNNs, to solve problems in not discrete domains. To the best of the authors' knowledge, our approach to continuous convolution has not been explored in literature yet.

The main novelties of this work rely on:
\begin{itemize}
    \item Building a new framework, based on continuous filters, for working with unstructured data (continuous filter).
    \item Defining a neat \reviewerB{analogy} between continuous (transposed) convolution and state of the art discrete (transposed) convolution in CNNs.
    \item Apply continuous convolutional layers in a CNN with partially-completed \reviewerB{input}.
    \item Exploiting general strategies to work with continuous convolutional autoencoders for dimensionality reduction and system output predictions at unseen time steps.
    \end{itemize}
\reviewerA{All this, we highlight, preserving the features of the standard CNNs, which make such an approach effective even dealing with large datasets.}
The present contribution is organised as follows: in Section \ref{sec2}, a small review of deep learning architectures useful for later analysis is done, as well as introducing the continuous filter for one-dimensional and multi-dimensional channels. In the same Section, we introduce the main idea to perform transposed continuous convolution. Section \ref{sec3} is focused on numerical results. First, we validate the proposed methodology on a discrete domain problem using a continuous CNN and compare it with its discrete representation. Second, we show that continuous convolution can also work with partially-completed images. Last, we present different deep learning architectures using continuous filters to solve the step Navier Stokes problem, and the  multiphase problem. Finally, conclusions follow in Section \ref{sec4}.

\section{Methodology}\label{sec2}
This Section focuses on the various methodologies we rely for building the continuous filter, as well as the introduction of the framework. First of all, we will describe briefly the feed-forward neural network and the discrete filter for a CNN in Section \ref{feed-forward} and Section \ref{discrete-filter} respectively. As already mentioned in Section \ref{sec1}, one of the main novelty of the work is building a new framework based on continuous convolution. Hence, Section \ref{continous-filter} concerns the introduction of our framework in different settings: single channel, multiple channel and transposed convolution using the continuous filter.

    \subsection{Feed-Forward Neural Network}\label{feed-forward}
Feed-forward Neural Network, or \emph{multi-layer perceptron} (MLP), is the most basic, yet one of the most important, building block of most current deep learning architectures \cite{Goodfellow-et-al-2016, fine2006feedforward, calin2020deep}. Widely used in deep learning, MLPs have the ability to approximate any continuous function due to the universal approximation theorem \cite{mlp-approximator, cybenko1989approximation, LESHNO1993861}.
\begin{figure}[!ht]
    \centering
\resizebox{.9\textwidth}{!}{%
    \tikzset{%
      every neuron input/.style={
        circle,
        draw,
        minimum size=0.7cm,
        fill=green!50
      },
      every neuron input2/.style={
        circle,
       draw,
        minimum size=0.7cm,
        fill=yellow!50
      },
       every neuron hidden/.style={
        circle,
        draw,
        minimum size=0.7cm,
        fill=blue!50
      },
       every neuron output/.style={
        circle,
        draw,
       minimum size=0.7cm,
        fill=red!50
      },
      neuron missing/.style={
        draw=none,
        fill=none,
        scale=4,
        text height=0.333cm,
        execute at begin node=\color{black}$\vdots$
      },
       layer missing/.style={
        draw=none,
        scale=4,
        text height=0.333cm,
        execute at begin node=\color{black}$\dots$
      },
    }
    \begin{tikzpicture}[x=1.5cm, y=1.5cm, >=stealth, font=\small, scale=0.9]
    \foreach \i [count=\y] in {1, 2, missing, 3}
    \node [every neuron input2/.try, neuron \i/.try] (input-\i) at (0,2.5-\y) {};
    \foreach \m [count=\y] in {1, 2, missing,3}
      \node [every neuron hidden/.try, neuron \m/.try ] (hidden1-\m) at (2,2.5-\y) {};
    \foreach \m [count=\y] in {1, missing, 2, 3}
      \node [every neuron hidden/.try, neuron \m/.try ] (hidden2-\m) at (4,2.5-\y) {};
    \foreach \m [count=\y] in {1,missing,2}
      \node [every neuron output/.try, neuron \m/.try ] (output-\m) at (6,1.9-\y) {};
    \draw [<-] (input-3) -- ++(-1,0)
        node [above, midway] {$x_{n_{\text{in}}}$};
    \draw [<-] (input-1) -- ++(-1,0)
        node [above, midway] {$x_1$};
    \draw [<-] (input-2) -- ++(-1,0)
        node [above, midway] {$x_2$};
    \foreach \l [count=\i] in {1,{n_{\text{out}}}}
      \draw [->] (output-\i) -- ++(1,0)
        node [above, midway] {$\hat{y}_{\l}$};
    \foreach \i in {1,...,3}
      \foreach \j in {1,...,3}
        \draw [->] (input-\i) -- (hidden1-\j);
    \foreach \i in {1,...,3}
      \foreach \j in {1,...,3}
        \draw [->] (hidden1-\i) -- (hidden2-\j);
    \foreach \i in {1,...,3}
      \foreach \j in {1,...,2}
        \draw [->] (hidden2-\i) -- (output-\j);
    \foreach \l [count=\x from 0] in {Input \\ layer , Hidden \\ layer 1}
    \node [align=center, above] at (\x*2,2) {\l};
    \node [align=center, above] at (4,2) {Hidden \\ layer 2};
    \node [align=center, above] at (6,2) {Output \\ layer };
    \end{tikzpicture}
}
\caption{Schematic structure of Feed-Forward Neural Network.\label{fig:FNN}}
\end{figure}
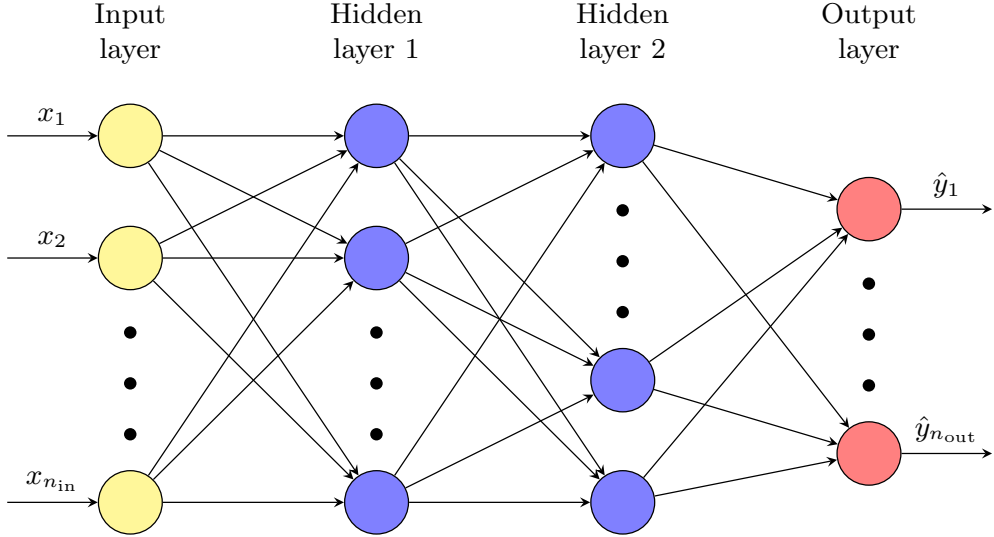
More technically, given an input vector $\mathbf{x} \in \mathbb{R}^{n_\text{in}}$ and a function to approximate $\phi : \mathbb{R}^{n_{\rm{in}}} \rightarrow \mathbb{R}^{n_\text{out}}$; the MLP approximation is done using a parameterised function class $\mathcal{F} = \{f_{\bm{\theta} \in \Theta} \}$, where $\bm{\theta}$ are trainable parameters of the network, belonging to the parameters' space $\Theta$. A MLP can be represented as a directed acyclic graph, as depicted in Figure \ref{fig:FNN}. In particular, it is composed by an input layer, an output layer and a certain number of hidden layers, where the processing units of network, called \textit{neurons}, perform the computation. Each layer $i$, with $i \in 0, \dots, M$, can be thought as a function $f^{(i)}$ belonging to $\mathcal{F}$, and the overall network function is given by the layers' composition~\cite{deisenroth2020mathematics}:
\begin{equation}\label{eqn:net-composition}
    f = f^{(M)} \circ f^{(M-1)} \circ \cdots \circ f^{(1)} \circ f^{(0)}.
\end{equation}

Hence, a single layer $i$, is a function $f^{(i)} : \mathbb{R}^{n_i} \rightarrow \mathbb{R}^{n_{i+1}} $, where $n_i$, represents the number of neurons in layer $i$, with $n_0 = n_{\rm{in}}$ and $n_{M+1} = n_{\rm{out}}$. Each layer $i$ is composed by $\theta_i = (\mathbf{w}_{(i)}, \mathbf{b}_{(i)})$ parameters, where $\mathbf{w}_{(i)}$ is a real matrix $n_{i+1}\times n_i$, called \textit{weight matrix}, and $\mathbf{b}_{(i)}$ is a real vector of dimension $n_{i+1}$, called \textit{bias}. The output vector $\mathbf{h}_{(i+1)}$ of layer $i$, corresponding to input vector of layer $i+1$ (except for the output layer), is then calculated using:
\begin{equation}\label{eqn:single-layer}
    \mathbf{h}_{(i+1)} = f^{(i)} (\mathbf{h}_{(i)} \mid \theta_i) = \delta^{(i)} (\mathbf{w}_{(i)}\cdot\mathbf{h}_{(i)}  + \mathbf{b}_{(i)}),
\end{equation}
where $\mathbf{h}_{(0)} = \mathbf{x}$, and $\mathbf{h}_{(M+1)} = \mathbf{\hat{y}}$ is the output of the network. The function $\delta^{(i)} : \mathbb{R}^{n_i} \rightarrow \mathbb{R}^{n_{i+1}} $ is called \textit{activation}, introducing non-linearity through the network; common choices are represented by the ReLU function, the sigmoid, the logistic function or the radial activation functions. By using Equation \ref{eqn:net-composition} and Equation \ref{eqn:single-layer}, one can express mathematically a MLP architecture.\\


During the \textit{training process}, in which a data-set $\mathcal{D} = \{(\mathbf{x}_i, \phi(\mathbf{x})_i )\}_{i=1}^n$ composed by $n$ observation is fed into the network, the MLP parameters $\bm{\theta}$ are modified in order to minimize a loss function $\mathcal{L}(\bm{\theta} \mid \mathcal{D},\,f)$. The choice of the loss function depends on the specific problem of application~\cite{Goodfellow-et-al-2016, book_NN, calin2020deep}. Hence, the learning phase can be summarised mathematically as:
\begin{equation}
     \min_{\bm{\theta}}
     \big\{\mathcal{L}(\bm{\theta} \mid \mathcal{D},\, f)\big\}.
\end{equation}

In practice, to solve the minimization problem, different optimization algorithms  based on back-propagation can be used, see \cite{rojas1996backpropagation, minimization, zaki2020data} for further reference. The optimization phase is done in multiple \textit{training epochs}, i.e. a complete repetition of the parameter update involving the complete training data-set $\mathcal{D}$.
    \subsection{Discrete filter in Convolutional Neural Networks}\label{discrete-filter}
\begin{figure}[thb]
    \centering
    \resizebox{.9\linewidth}{!}{%
    \begin{tikzpicture}[scale=1.0]

  \matrix [nodes=draw,column sep=-0.2mm, minimum size=6mm]
  {
    \node {0}; & \node{1}; & \node {1}; & \node{1}; & \node{0}; & \node{0}; & \node{0}; \\
    \node {0}; & \node{0}; & \node {1}; & \node{1}; & \node{1}; & \node{0}; & \node{0}; \\
    \node {0}; & \node{0}; & \node {0}; & \node{1}; & \node{1}; & \node{1}; & \node{0}; \\
    \node {0}; & \node{0}; & \node {0}; & \node{1}; & \node{1}; & \node{0}; & \node{0}; \\
    \node {0}; & \node{0}; & \node {1}; & \node{1}; & \node{0}; & \node{0}; & \node{0}; \\
    \node {0}; & \node{1}; & \node {1}; & \node{0}; & \node{0}; 
    & \node{0}; & \node{0}; \\
    \node {1}; & \node{1}; & \node {0}; & \node{0}; & \node{0}; 
    & \node{0}; & \node{0}; \\
  };

  \coordinate (A) at (-0.3,0.3);
  \coordinate (B) at (1.5,0.3);
  \coordinate (C) at (1.5,2.12);
  \coordinate (D) at (-0.3,2.12);
  \fill[blue, opacity=0.2] (A)--(B)--(C)--(D)--cycle;
  \begin{scope}[shift={(3.3,0)}]
    \node[] at (0,0) {\Large $\ast$};
  \end{scope}[shift={(2.5,0)}]

  \begin{scope}[shift={(5,0)}]

    \matrix [nodes=draw,column sep=-0.2mm, minimum size=6mm]
    {
      \node{1};  & \node{0};   & \node{1};  \\
      \node{0};  & \node{1};   & \node{0};  \\
      \node{1}; & \node{0}; & \node{1}; \\
    };
    \coordinate (A1) at (-0.9,-0.9);
    \coordinate (B1) at (0.93,-0.9);
    \coordinate (C1) at (0.93,0.92);
    \coordinate (D1) at (-0.9,0.92);
    \fill[blue, opacity=0.2] (A1)--(B1)--(C1)--(D1)--cycle;
  \end{scope}

  \begin{scope}[shift={(6.6,0)}]
    \node[] at (0,0) {\Large $=$};
  \end{scope}[shift={(2.5,0)}]

  \begin{scope}[shift={(9,0)}]

    \matrix [nodes=draw,column sep=-0.2mm, minimum size=6mm]
    {
      \node{1};  & \node{4};   & \node{3}; & \node{4}; & \node{1};  \\
      \node{l};  & \node{2};   & \node{4}; & \node{3}; & \node{3};  \\
      \node{1}; & \node{2}; & \node{3}; & \node{4} ; & \node{1};  \\
      \node{1}; & \node{3}; & \node{3}; & \node{1} ; & \node{1};  \\
      \node{3}; & \node{3}; & \node{1}; & \node{1} ; & \node{0};  \\
    };
    \coordinate (A2) at (0.3,0.9);
    \coordinate (B2) at (0.91,0.9);
    \coordinate (C2) at (0.91,1.507);
    \coordinate (D2) at (0.3,1.507);
    \fill[blue, opacity=0.2] (A2)--(B2)--(C2)--(D2)--cycle;
  \end{scope}

\end{tikzpicture}
    }
    \caption{Discrete convolution operation on one dimensional tensor.}
    \label{fig:conv}
\end{figure}
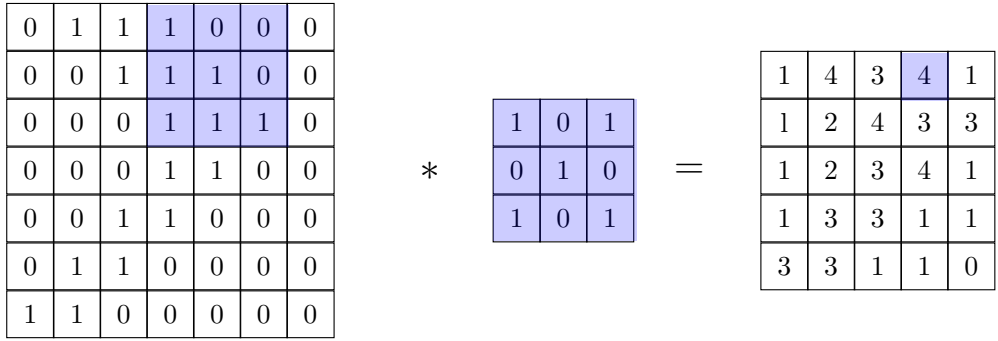

Convolutional Neural Network (CNN) is a class of deep learning architectures, vastly applied in computer vision \cite{cnn_image_class, krizhevsky2012imagenet, shanmugamani2018deep, jiang2019deep}. Over the past years, different CNN architectures have been presented, for instance AlexNet~\cite{imagenet}, ResNet~\cite{resnet}, Inception~\cite{inception}, VGGNet~\cite{vggnet}. 
Differently from MLPs, in which affine transformations are performed for learning, a convolutional layer actually performs the convolution of the input data $\mathcal I$ and the so called convolutive filter $\mathcal K$, such that
\begin{equation}
(\mathcal I\ast \mathcal K)(\mathbf{x})=\int_{-\infty}^{\infty} \mathcal I(\mathbf{x} + \boldsymbol{\tau})\mathcal K(\boldsymbol{\tau})d\boldsymbol{\tau}.
\end{equation}
CNNs perform such convolution\footnote{In many deep learning implementations the term convolution indicates what is known in mathematics as cross-correlation~\cite{Goodfellow-et-al-2016}. In this text, the term convolution will be used to indicate cross-correlation, thus adapting to the deep learning community convention.} in a discrete setting, using a tensorial representation of the two functions $\mathcal I$ and $\mathcal K$ instead of their continuous formulation. 
Thus, discrete correlation is computed as $(\mathcal I\ast \mathcal K)(\mathbf x) = \textstyle \sum_{\boldsymbol{\tau}=-\infty}^{\infty} \mathcal I(\mathbf {x}+\boldsymbol{\tau})\mathcal K(\boldsymbol{\tau})$, with $\mathbf x, \boldsymbol{\tau} \in \mathbb Z^d$ (with $d$ dimensions), where the latter infinite summation can be truncated by discarding the null products. In this way, it is not necessary to know the original function $\mathcal I$, but its evaluation at discrete coordinates.
In this context, the filter $\mathcal K$ can be represented as the tensor $\mathbf K \in \mathbb R^{N_1\times\dots\times N_d}$ such that the element $\mathbf K_{i_1, \dots, i_d}  \equiv \mathcal K(i_1, \dots, i_d)$ with $i_j \in \{1, \dots, N_j\}, \forall j \in \{1, \dots, d\}$. 
Applying a similar representation also for the input, the convolution results in the sum of the element-wise multiplication between input and filter, as sketched in Figure~\ref{fig:conv}. The convolution is of course repeated for all the input components, by moving the filter across the input in a regularized fashion~\cite{Goodfellow-et-al-2016, calin2020deep}.

The filter components (the so-called weights) represent the trainable parameters of the convolutional layer, which are tuned during the training phase.
In general, convolution reduces the size of a (multidimensional) array, performing \textit{downsampling}. 
Conversely, the opposite transformation to downsampling, called \textit{upsampling}, used by many deep learning architectures, e.g. autoencoders, uses \emph{transposed convolution}. The interested reader might refer to \cite{convolution_review, zhang2021dive} for more information regarding discrete (transposed) convolution.
    \subsection{Continuous filter}\label{continous-filter}
In contrast to discrete convolution as described in the previous Section, continuous two-dimensional convolution is mathematically defined as:
\begin{equation}\label{continous_convolution}
    \mathcal{I}_{\rm{out}}(x, y) = \int_{\mathcal{X}}\int_{\mathcal{Y}}  \mathcal{I}(x + \tau_x, y + \tau_y) \cdot \mathcal{K}(\tau_x, \tau_y) d\tau_x d\tau_y,
\end{equation}
where $\mathcal{K} : \mathcal{X} \times \mathcal{Y} \rightarrow \mathbb{R}$ is the \textit{continuous filter} function, and $\mathcal{I} : \Omega \subset \mathbb{R}^2 \rightarrow \mathbb{R}$ is the input function. The continuous filter function is approximated using a MLP, thus trainable during the training phase. In order to maintain the parallelism with discrete convolution in CNNs, the definition adopted for continuous convolution differs from the mathematical one for which  $\mathcal{X} = \mathcal{Y} = \mathbb{R}$. In fact, the continuous filter presented is defined on a close domain, smaller than the input function domain, as in the case of the discrete filter. The integral in Equation \ref{continous_convolution} can be evaluated using different techniques from numerical analysis \cite{numerical_integration}. In our implementation, for simplicity, the double integral is approximated by a double sum on the grid nodes inside the filter domain.
Given $\{(x_i, y_i)\}_{i=1}^{n}$ points of the input function mapped on the $\mathcal{X} \times \mathcal{Y}$ filter domain, we approximate Equation \ref{continous_convolution} as:
\begin{equation}\label{continous_convolution_approx}
    \mathcal{I}_{\rm{out}}(\tilde{x}_i, \tilde{y}_i) = \sum_{{x_i}\in\mathcal{X}}\sum_{{y_i}\in\mathcal{Y}}  \mathcal{I}(x_i + \tau_x, y_i+ \tau_y) \cdot \mathcal{K}(x_i, y_i),
\end{equation}
where $(\tau_x, \tau_y) \in \mathcal{S}$, with $\mathcal{S}$ the set of available strides, corresponds to the current stride position of the filter, and $(\tilde{x}_i, \tilde{y}_i)$ points  are obtained by taking the centroid of the filter position mapped on the $\Omega$ domain. 
It is important to remark, that the discretization strategy for $\Omega$ is irrelevant for the filter usage, since it can work even for holed or not connected domains. Finally, the strides positions $\mathcal{S}$, constituting the different positions of the filter on $\Omega$, do not have to cover the whole $\Omega$ domain, thus representing a generalization of the discrete filter stride. 

Having described the overall mathematical idea behind one channel continuous convolutions, in Algorithm \ref{Algorithm_convolution} a possible simple implementation is reported. When implementing the continuous convolutional layer, we identified opportunities to optimize it (e.g., stacking mapped input to call just once the MLP or calling the mapping function on the entire data before entering in the \texttt{for} loop, allowing just one point search per forward pass). The pseudo-code in Algorithm \ref{Algorithm_convolution} does not report, for simplicity, these optimizations used in building the layer. Moreover, the mapping function, which is responsible for finding the points inside the filter, can be implemented via linear search or advanced algorithms based on efficient data structures. The usage of efficient data structures may decrease the overall time complexity of the algorithm, leading to an efficient implementation of the convolutional layer. Note that the implementation used in this article relies on linear search for the mapping function.

\begin{algorithm}
\caption{One channel continuous convolution algorithm}
\label{Algorithm_convolution}
\begin{algorithmic}
\Require $\mathcal{I}$ : input function  
\Require $\mathcal{K}$ : MLP approximation function
\Require $\mathcal{S}$ : set of stride positions
\Require $\Omega$ : domain discretization 
\State $\mathcal{I}_{\rm{out}} \gets$ \textbf{Initialize($\Omega$, $\mathcal{S}$)} \Comment{Initialize output coordinates}
\For{$(\tau_x, \tau_y) \in \mathcal{S}$}
%
%

   $(\mathbf{x}, \mathbf{y}) \gets$ \textbf{Map($\Omega$, $(\tau_x, \tau_y)$)} \Comment{$\Omega \to \mathcal X \times \mathcal Y$}
   
   $\mathbf{v}_{\mathcal{K}} \gets \mathcal{K}(\mathbf{x}, \mathbf{y})$ \Comment{MLP evaluation}
   
   $ \mathcal{I}_{\rm{out}} \gets$ \textbf{Convolve}($\mathcal{I}(\mathbf{x} + \tau_x, \mathbf{y} + \tau_y), \mathbf{v}_{\mathcal{K}})$ \Comment{Convolution}
\EndFor \\
\Return $\mathcal{I}_{\rm{out}}$
\end{algorithmic}
\end{algorithm}

The latter is easily extendable for multichannel convolution. Here two approaches can be adopted: multiple two-dimensional filters, or multiple three-dimensional filters. In the first case, the framework is  analogous to the one of the discrete convolution in multi dimensional channels, see Section \ref{discrete-filter}. We suggest to use this if the channels represent independent quantities (e.g., pressure and velocity in a fluid, or image channels). The second possibility is to define a three-dimensional filter, meaning a function $\mathcal{K} : \mathcal{X} \times \mathcal{Y} \times \mathcal{P} \rightarrow \mathbb{R}$, which takes as extra argument a channel parameter $p \in \mathcal{P}$. We suggest to use this if the channels represent correlated quantities (e.g., velocity along two directions in a fluid). While the multiple two-dimensional filters strategy needs $\rm{Output\, Channels}\times \rm{Input\, Channels}$ independent filters, i.e. neural networks; the multiple three-dimensional filters strategy only needs $\rm{Output\, Channels}$ independent filters, since the multichannel in the input is already handled by the extra dimension in the filter. In our implementation we considered and tested both strategies. 

\begin{figure}[thb]
\centering
\resizebox{\linewidth}{!}{
\begin{tikzpicture}[scale=1.0]

  \matrix [nodes=draw,column sep=-0.2mm, minimum size=6mm,label={[font=\large]above:input}]
  {
    \node {0}; & \node{1};  \\
    \node {2}; & \node{3}; \\
  };

  \coordinate (A) at (-0.6,0.0);
  \coordinate (B) at (0.0,-0.);
  \coordinate (C) at (0.0,0.6);
  \coordinate (D) at (-0.6,0.6);
  \fill[blue, opacity=0.2] (A)--(B)--(C)--(D)--cycle;
  
  \coordinate (A) at (+0.6,0.0);
  \coordinate (B) at (0.0,0.);
  \coordinate (C) at (0.0,-0.6);
  \coordinate (D) at (+0.6,-0.6);
  \fill[blue, opacity=0.4] (A)--(B)--(C)--(D)--cycle;

  \begin{scope}[shift={(1,0)}]
    \node[] at (0,0) {\Large $\ast$};
  \end{scope}[shift={(1,0)}]

  \begin{scope}[shift={(2,0)}]

    \matrix [nodes=draw,column sep=-0.2mm, minimum size=6mm,label={[font=\large, yshift=0.1cm]above:kernel}]
    {
      \node{0};  & \node{1};  \\
      \node{2};  & \node{3};  \\
    };

    \coordinate (A1) at (-0.6,-0.6);
    \coordinate (B1) at (0.6,-0.6);
    \coordinate (C1) at (0.6,0.6);
    \coordinate (D1) at (-0.6,0.6);
    \fill[blue, opacity=0.2] (A1)--(B1)--(C1)--(D1)--cycle;
  \end{scope}

  \begin{scope}[shift={(3,0)}]
    \node[] at (0,0) {\Large $=$};
  \end{scope}[shift={(3,0)}]

  \begin{scope}[shift={(4.3,0)}]

    \matrix [nodes=draw,column sep=-0.2mm, minimum size=6mm]
    {
      \node{0};  & \node{0};   & \node{};  \\
      \node{0};  & \node{0};   & \node{};  \\
      \node{}; & \node{}; & \node{}; \\
    };
    \coordinate (A1) at (-0.9,-0.3);
    \coordinate (B1) at (0.3,-0.3);
    \coordinate (C1) at (0.3,0.92);
    \coordinate (D1) at (-0.9,0.92);
    \fill[blue, opacity=0.2] (A1)--(B1)--(C1)--(D1)--cycle;
  \end{scope}
  
    \begin{scope}[shift={(5.6,0)}]
    \node[] at (0,0) {\large $+$};
  \end{scope}[shift={(5,0)}]

    \begin{scope}[shift={(6.9,0)}]

    \matrix [nodes=draw,column sep=-0.2mm, minimum size=6mm]
    {
      \node{};  & \node{0};   & \node{1};  \\
      \node{};  & \node{2};   & \node{3};  \\
      \node{}; & \node{}; & \node{}; \\
    };
  \end{scope}
  
  \begin{scope}[shift={(8.2,0)}]
    \node[] at (0,0) {\large $+$};
  \end{scope}[shift={(2.5,0)}]

    \begin{scope}[shift={(9.5,0)}]

    \matrix [nodes=draw,column sep=-0.2mm, minimum size=6mm]
    {
      \node{};  & \node{};   & \node{};  \\
      \node{0};  & \node{2};   & \node{};  \\
      \node{4}; & \node{6}; & \node{}; \\
    };
  \end{scope}
  
    \begin{scope}[shift={(10.8,0)}]
    \node[] at (0,0) {\large $+$};
  \end{scope}[shift={(2.5,0)}]
  
      \begin{scope}[shift={(12.1,0)}]

    \matrix [nodes=draw,column sep=-0.2mm, minimum size=6mm]
    {
      \node{};  & \node{};   & \node{};  \\
      \node{};  & \node{0};   & \node{3};  \\
      \node{}; & \node{6}; & \node{9}; \\
    };
    
    \coordinate (A1) at (0.9,0.3);
    \coordinate (B1) at (-0.3,0.3);
    \coordinate (C1) at (-0.3,-0.92);
    \coordinate (D1) at (0.9,-0.92);
    \fill[blue, opacity=0.4] (A1)--(B1)--(C1)--(D1)--cycle;
  \end{scope}
  
      \begin{scope}[shift={(13.4,0)}]
    \node[] at (0,0) {\Large $=$};
  \end{scope}[shift={(2.5,0)}]
  
     \begin{scope}[shift={(14.7,0)}]

    \matrix [nodes=draw,column sep=-0.2mm, minimum size=6mm,label={[font=\large]above:output}]
    {
      \node{0};  & \node{0};   & \node{1};  \\
      \node{0};  & \node{4};   & \node{6};  \\
      \node{4}; & \node{12}; & \node{9}; \\    };
  \end{scope}
  
\end{tikzpicture}
}
\caption{Example of transposed convolution in discrete setting.}\label{fig:transp}
\end{figure}
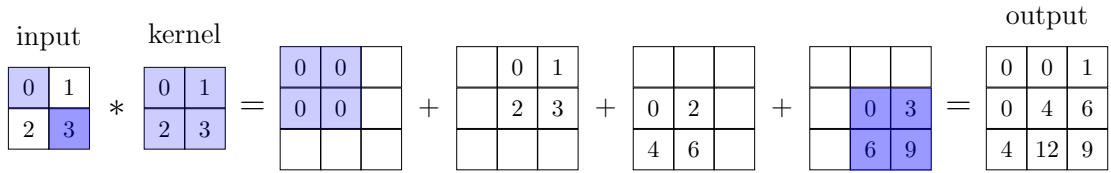

Finally, we highlight that such filter allows to trivially extend the transposed convolution in the continuous setting. In several architectures, like the autoencoder employed in the numerical experiments of this contribution, it is indeed necessary to upsample the output of a given layer. With discrete filters, such operation is obtained by simply multiplying any elements in the input tensor by all the elements of the filter in a element-wise fashion. Figure~\ref{fig:transp} sketches a simple example of transposed convolution with 2D tensors. In case of overlapping elements in the output --- such condition depends by the stride used in the transposed convolution --- these elements are typically summed, but also other choices can be employed, like averaging.
In the continuous counterpart, the transposed operation is semantically the same, but the elements within the filter are not anymore finite. It is necessary indeed, in this latter case, to select the points within the filter domain, sample the kernel function at those points and finally \reviewerA{weight} them with the elements of the input tensor to produce the output.

\section{Numerical Results}\label{sec3}
In this Section different experiments and their results, using our continuous convolutional filter, are presented. In Section \ref{mnist_prob} a comparison between the state-of-the-art discrete filter and the continuous filter is done, while in Section \ref{partially_compelted} the advantages of using a continuous filter for partially-complete images are shown. Following, in Section \ref{NS_prob} we validate the usage of transposed convolution using a Navier Stokes problem. Finally, in Section \ref{liquid_gas_prod}, the continuous filter is used to build an autoencoder architecture for reconstructing not seen time snapshot for the multiphase problem.
\subsection{Software}
In order to implement and construct the continuous convolutional filter, as well as performing all the experiments in this Section, we employed PyTorch~\cite{pytorch} due to its versatility and its wide use in the deep learning community. Moreover, the open source C++ finite volume library OpenFOAM~\cite{of} is used for the mesh creation for the problems presented in Section~\ref{NS_prob} and Section~\ref{liquid_gas_prod}, and the FEniCSx \cite{LoggMardalEtAl2012} for solving the Navier Stokes problem in Section~\ref{NS_prob} using finite elements method. Finally, the EZyRB Python library \cite{Demo_EZyRB_Easy_Reduced_2018} is used for performing a comparison with proper orthogonal decomposition in Section~\ref{liquid_gas_prod}.

\subsection{Convolutional Neural Networks comparison on MNIST dataset}\label{mnist_prob}
The first problem presented for validating the methodology is the classification task in a supervised learning setting. Training and testing is done on the MNIST dataset \cite{mnist_dataset}, as it represents a standard benchmark for classification. In particular we compare two CNNs, differing only for the first convolutional layer in which discrete and continuous filters are interchanged. The MNIST dataset used during test is composed by $60000$ training images, and $10000$ testing images. The images are saved as $1\times28\times28$ matrix $I$, where the first dimension represents the number of channels, while second and third dimension represent height and width respectively. When passing through the continuous filter the image is transformed using a \reviewerA{bed of nail} representation
\begin{equation}\label{bend_nails}
    I_\text{cont}(x, y) = \sum_{i,j} 
\delta(x - i, y- j)I[i, j],
\end{equation}
where the sum is done on the matrix indices, as also done in \cite{shocher2020discrete}. This representation allows to treat a matrix as a continuous function, thus permitting the use of the continuous filter. The opposite transformation is done by arranging the function value in a matrix whose pixel positions are given by the $(x, y)$ coordinates of the function $I_\text{cont}$.

\begin{figure}[t!]
    \centering
    \includegraphics[width = 0.95\linewidth]{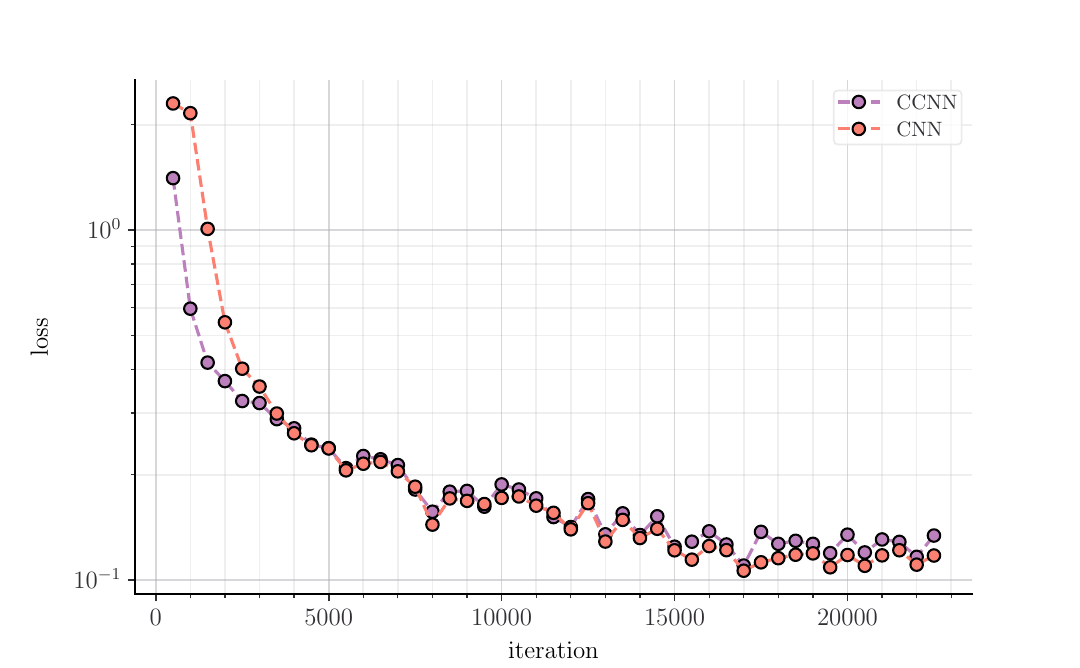}
    \caption{\reviewerA{Comparison of CCNN and CNN with respect to loss value and the number of iterations.}}
    \label{loss_plot}
\end{figure}

Due to construction, the continuous filter is expected to perform equivalently to a discrete filter with structured data, thus classification on MNIST can be used as a validation methodology for the continuous filter framework. The general network used in training and testing is composed by two convolutional layers followed by three fully-connected layers of size $150,~24,$ and $10$ respectively, with hyperbolic tangent activation after the first two layers. The first convolutional layer has filter size and stride both equal to $4$, with one input and output channel, and zero padding. The second convolutional layer has filter size and stride both equal to $1$, with one input channel and four output channels, and zero padding. For the continuous filter we used a MLP with two layers of size $12, 12$ with ReLU activation for the inner network. Through the experiments only the first convolutional layer is modified in the general network by changing discrete with continuous filters (and vice-versa), but without changing any hyper-parameter, e.g. stride, filter size or padding. Let us indicate with CNN the general network using a discrete filter in the first layer, and CCNN the continuous version of the CNN, i.e. the one in which the first convolutional layer discrete filter is replaced with a continuous filter. Both networks are trained, minimizing the cross entropy loss \cite{Goodfellow-et-al-2016}, for $22500$ iterations on the training set using the Stochastic Gradient Descent optimizer \cite{optimizers} with learning rate and momentum set to $0.001$ and $0.9$ respectively, and batch size equal to $8$.

In Figure \ref{loss_plot} the training loss for the two networks is reported at different iterations. The result shows a faster convergence of the CCNN network with respect to the CNN network. Nevertheless, at plateau, both networks reach similar level of training loss, as expected. Furthermore, both networks have a similar decay shape of the loss, evidencing that the training process on a continuous filter using discrete data is comparable to a discrete filter. 
\begin{table}[thb!]
    \centering
\caption{\reviewerB{Results obtained for CNN and CCNN networks trained on MNIST dataset. Train time and test time reported in seconds.}}
    \label{tab:results_mnist}
\begin{tabular}{@{}cccccc@{}}
    \toprule 
    \textbf{Network} & \textbf{Train Accuracy}  & \textbf{Test Accuracy}  & \textbf{Size} & \textbf{Train Time}  &
    \textbf{Test Time} \\
    \midrule
    CCNN  &  $96.6\%$ & $96.1\%$ & $33593$ & $630$ & $21.1$\\
    CNN & $96.8\%$ & $96.6\%$  & $33449$ & $40$ & $1.1$ \\
    \bottomrule
\end{tabular}
\end{table}

In Table \ref{tab:results_mnist} the training accuracy, the testing accuracy, the size of the networks used during training and the training time are reported. 
The accuracy is defined as the number of correctly classified predictions divided by the total number of predictions. It is possible to note from the table that the training and testing accuracy is approximately the same between the two networks, thus confirming heuristically the hypothesis that both networks behave the same on \reviewerA{structured data}. The discrete filter tends to have a small percentage advantage $0.2\%$ on training data, and $0.5\%$ on test data over the continuous filter. Nevertheless, fine-tuning the inner network for the continuous filter might increase the test accuracy as well. Furthermore, we stress that the true power of continuous filters is employed when unstructured data are used, and the test only wants to confirm the validity of the methodology. The CCNN network presents a small overhead in terms of the size of the network due to the MLP used for approximating the continuous filter function. Nevertheless, the sizes of the two networks are still comparable in order of magnitude. Finally, due to expensive routines, e.g. the mapping function, the CCNN is slower compared to a standard CNN, resulting in higher training time. To summarize, in terms of accuracy using a CNN or a CCNN does not affect the overall performance. Even so, CCNN can be slightly slower than normal CNN due to continuous filter architecture construction, but CCNN can work with unstructured domains. 
In the following experiments, the power of continuous filter will be shown by working with unstructured domains.
\subsubsection{Partially-completed images}\label{partially_compelted}
Continuous convolutional neural networks have the ability to work also with partially completed images, unlike normal convolutional neural networks. Recalling the \reviewerA{bed of nails} representation reported in Equation \ref{bend_nails}, some information can be removed from the image $I$ by avoiding some indices $(i^*, j^*)$ into the summation. This procedure corresponds to not evaluating the continuous representation of $I$ in $(x = i^*, y = j^*)$, thus removing information from the image. Notice that the $(i^*, j^*)$ positions are not filled with zeros, but they are actually not considered by the continuous filter. This procedure can not be done on a regular discrete filter since it would require to remove the $(i^*, j^*)$ entry of the matrix $I$, thus losing the structure in the data, and making not possible to perform matrix operations. On the contrary, by using a continuous representation of $I$, removing pixels corresponds to not evaluating the continuous function on them, and the filter can still be used since it can deal with unstructured data by construction.\\
The second problem presented deals with the classification task on a supervised learning setting using partially-completed images. In order to understand if a continuous filter can work with missing pixels images, i.e. the filter is able to generalize with loss of information, a simple two-layer network is used. In the first layer a continuous filter of size and stride both equal to $4$ is posed, followed by the ReLU activation function and a linear layer (single MLP layer) of size $10$ used to classify the output of the convolution. By adopting this simple network it is possible to understand how a single continuous filter is able to classify images as the pixel information reduces. Indeed, the network is not built for performance in accuracy, rather it is used to show that the continuous filter can work and achieve effective results also with unstructured missing pixels images. Training and testing is done on the MNIST dataset presented in the previous Section. The network is trained, minimizing the cross entropy loss for $22500$ iterations on the training set using the Stochastic Gradient Descent optimizer with learning rate set to 0.001, and batch size equal to 8. 

\begin{figure}[t!]
    \centering
    \includegraphics[width = 0.95\linewidth]{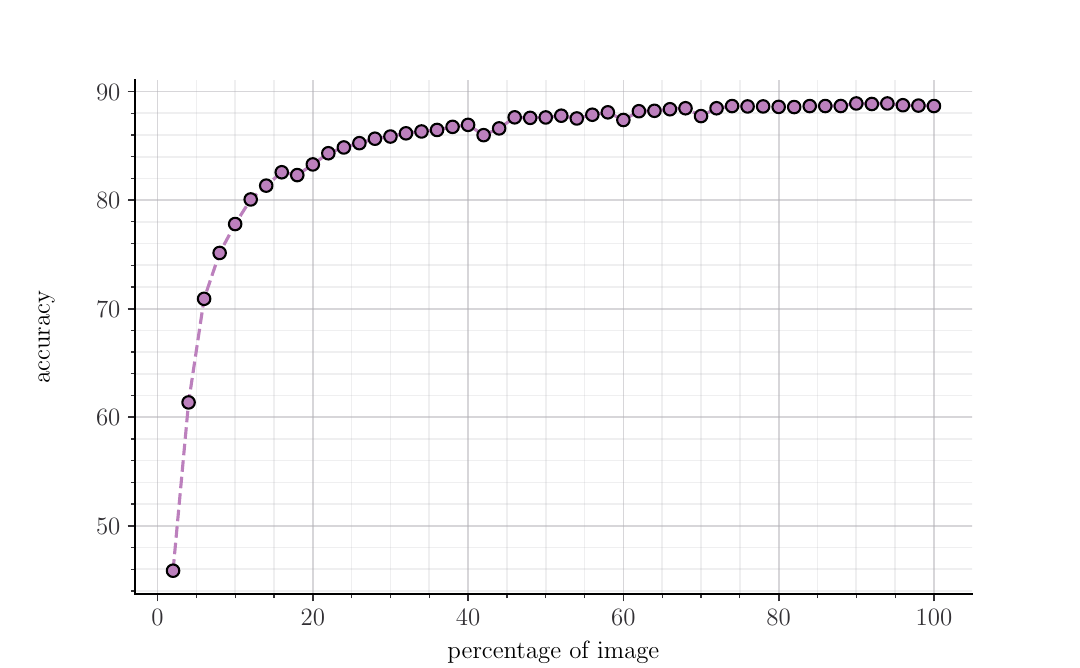}
    \caption{Variation in testing accuracy at difference percentage of information on input images. }
    \label{mnist_res_pic}
\end{figure}

Let us define $\mathcal{P} \in [0, 100]$ the percentage of information in the image, i.e. the percentage of pixels of the original image. The pixels are removed using a uniform two dimensional distribution, selecting the $(x = i^*, y = j^*)$ coordinates where not evaluating the continuous representation of $I$. The experiment is performed by training, for each different percentage in $\mathcal{P}$, the proposed network\footnote{By decreasing the number of informative pixels, the output size of the continuous convolutional filter decreases as well. Hence, the input of following linear layer is different for different percentage of images. Nevertheless, the overall structure of the network remains the same.}, and then validating the network on the testing data. In Figure \ref{mnist_res_pic} the results of the experiment are depicted, where the network accuracy is reported as a function of the percentage of image. The results show that the network is able to learn with missing value pixels. Starting from the baseline given by a percentage of $100$, the network is robust in learning, obtaining comparable results to the baseline till roughly a percentage of $75$. As the percentage decreases the accuracy also decreases, which is expected since information is hidden to the network. Nevertheless, with an image percentage of $20$ the network decreases in accuracy only for a small $8\%$ value, before a fast drop in accuracy due to little information fed into the network. Overall, it is evident that by using missing value images the continuous filter is able to generalize and learn, due to its ability of learning using unstructured domains. 
\subsection{Navier Stokes problem}\label{NS_prob}
\begin{figure}[thb]
    \centering
    \includegraphics[width=0.95\linewidth]{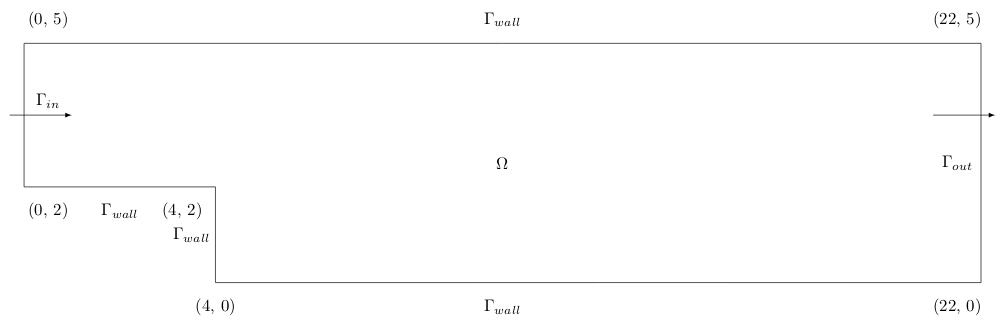}
    \caption{The domain for the Navier Stokes experiment.}
    \label{fig:ns_domain}
\end{figure}

The second experiment shows the capability of the continuous filter in a unsupervised learning framework using autoencoders \cite{autoencoders}. In detail, a typical CFD parametric benchmark is considered, where Navier Stokes equations are used to describe the flow in a back-step domain (Figure~\ref{fig:ns_domain}).
More specifically, equations and boundary conditions are set as follows:
\begin{equation}
\begin{array}{rll}
\nu \Delta \mathbf{u} + (\mathbf{u} \cdot \nabla)\mathbf{u} + \nabla p &= 0\quad\quad\quad&\text{in}\,\Omega,\\
\nabla \cdot \mathbf{u} &= 0\quad\quad\quad&\text{in}\,\Omega,\\
\mathbf{u} &=\mu\bigl\{\frac{1}{2.25}(x_1-2)(5-x_1),0\bigr\} \quad\quad\quad&\text{on}\,\Gamma_\text{in},\\
\mathbf{u} &= 0\quad\quad\quad&\text{on}\,\Gamma_\text{wall},\\
\nu \frac{\partial \mathbf{u}}{\partial \mathbf{n}} - p\mathbf{n} &= 0\quad\quad\quad&\text{on}\,\Gamma_\text{out},\\
\end{array}
\label{eq:stokes}
\end{equation}
where $\mathbf{u} \equiv \mathbf{u}(\mathbf{x}, \mu)$ and $p \equiv p(\mathbf{x}, \mu)$ are the spatial and parameter dependent velocity and normalized pressure fields, respectively, with $\mathbf{x} = (x_0, x_1) \in \Omega$ and $\mu \in [1, 80] \subset \mathbb R$. The letter $\mathbf{n}$ denotes the boundary normals, whereas the symbol $\nu$ indicates the kinematic viscosity, here imposed to $1.0$. Thus, we are imposing the so-called no-slip condition on the walls of the domain, a parametric horizontal velocity profile at the inflow and the directional do-nothing condition in the outflow.
To numerically solve such (nonlinear) problem, we have discretized the domain by means of a non-overlapping triangulation, onto which we compute the weak solution by exploiting a finite element framework.
Regarding the test functions, we have chosen a second order polynomial for the velocity space and a first order one for the pressure space, i.e. (P2, P1). 
The (parameter dependent) solutions belong then to an irregular discrete space, forbidding to apply standard CNNs to this kind of data. Thanks to the continuous filter proposed in the present work, we are able to perform convolution also in this setting, avoiding also post-processing steps --- e.g. projecting or interpolating the solutions on a regular grid.
\begin{figure}[th!]
    \centering
    \includegraphics[width=0.95\linewidth]{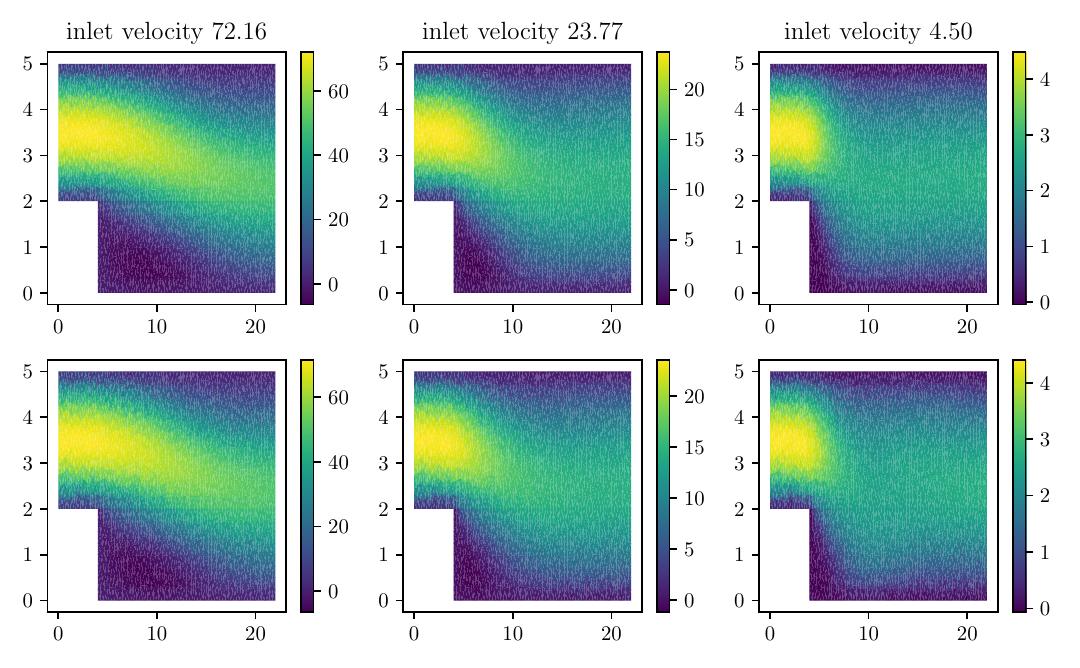}
    \caption{\reviewerB{Samples of CCAE reconstructed input (below) compared to the originals (above)}}.
    \label{ae_NS}
\end{figure}
An autoencoder (AE) is a type of deep learning architecture, which uses a MLP to encode the input into a lower dimensional meaningful representation, and then decode it back to obtain a reconstructed input as similar as possible to the original one. Convolutional autoencoder (CAE) is a type of AE where convolutional layers are used. In particular, CAE have been widely applied in fluid dynamics \cite{romor_ae, fluid_ae} as reduced order methods. However, as previously mentioned, pre-processing and post-processing steps are needed since discrete filters do not work with unstructured domains. Therefore, a continuous convolutional autoencoder (CCAE) is presented and used to extrapolate the latent representation of the Navier Stokes solution, without the need to pre-process or post-process it, thus keeping the unstructured domain representation.
In particular the proposed autoencoder is composed by an encoder, represented by a single channel continuous convolutional layer followed by a single fully-connected layer of size $90$ (latent representation size) and GELU \cite{gelu} activation function after it; and a decoder, represented by a fully-connected layer of size $840$ followed by a single channel transposed continuous convolutional layer. The (transposed) continuous convolutional layers have filter size and stride both equal to $[0.75, 0.18]$, while the inner network is approximated using a MLP of size $40, 40$ and GELU activation function. 
The CCAE is trained, minimizing the $l_1$ loss for $150$ epochs using the Adam optimizer \cite{adam} with learning rate set to $0.001$. The training data and testing data are divided by a $20-80\%$ rule on the total data set which is composed of $500$ solution samples at different inlet velocities of the fluid. Notice that the choice of a small percentage of data in the training set is made to guarantee that the network can generalize easily even with a low number of samples used during training.

Overall, the CCAE has an $l_2$ percentage error of $3.9\%$ both on training and testing. In Figure \ref{ae_NS} samples of reconstructed solutions using the CCAE are reported for different inlet velocities. The reconstructed samples show that the network has correctly generalized to unseen solutions, thus it can correctly extrapolate the latent dimension of them.
\begin{figure}[!t]
    \centering
    \includegraphics[width = 0.95\linewidth]{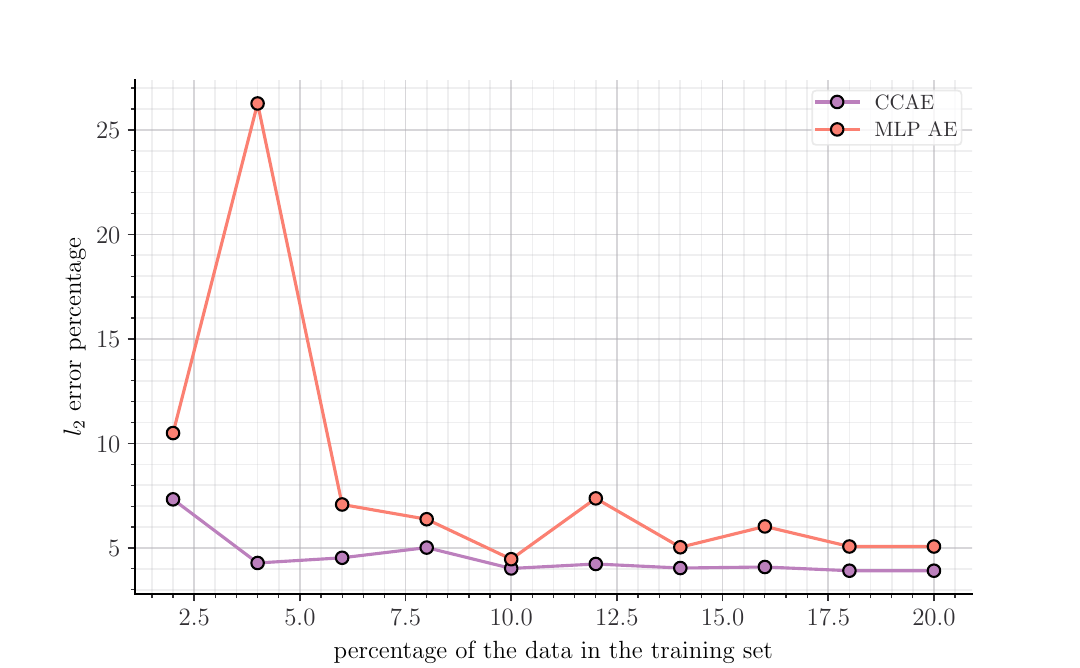}
    \caption{Variation in test error for different percentage of data in the training set.}
    \label{AEvsCCAE}
\end{figure}
A further meaningful test is to compare the CCAE with a simple AE that only uses a MLP for encoding and decoding. In fact, a MLP preserves the structure in the data as CCAE, avoiding pre- and post-processing. Nevertheless, we show that the MLP AE is characterized by a higher error rate than the CCAE. The test is done using the same CCAE structure defined above, while the MLP AE network is obtained by removing the continuous convolutional layer and the transposed continuous convolutional layer, thus it is represented by a MLP of input and output size of $1639$ (dimension of the Navier--Stokes solution vector), and hidden size of $90$ with GELU activation function after the first layer. Both networks are trained minimizing the $l_1$ loss, using the Adam optimizer with learning rate set to $0.001$. In order to see the ability to correctly learn a meaningful latent representation of the solution vector, both networks are trained multiple times. For each training, a different percentage of training data with respect to the overall data set size is used, while testing is done on the remaining solution samples. Hence, this test allows to compare the ability of both networks to generalize to unseen solutions as the number of data in training increases. The test is repeated $5$ times with different parameters initialization for both networks, and results are averaged to reduce noise.
The results of the experiment (Figure~\ref{AEvsCCAE}) highlight that both networks tend to reduce the overall test error as the number of training data increases, as expected. However, the CCAE tends to outperform the MLP AE constantly for different percentages of training data, which is shown by a lower $l_2$ error curve.
\subsection{Multiphase state problem}\label{liquid_gas_prod}
The last problem presented shows how a continuous filter can be used in a more advanced deep learning framework. Specifically, we present a deep learning architecture for performing inference on unseen snapshots in a computational fluid dynamics simulation using the multiphase problem. This problem is represented by two fluids occupying a certain fraction of the volume domain separated by a sharp interface. The two fluids evolve in time resulting in a wave moving in a specific direction. In order to perform the mesh we followed \cite{Papapicco_2022}, which gives also a mathematical formulation of the problem using the unsteady Navier Stokes equations. 
The mesh contains $18750$ grid points for a single snapshot, while $200$ different time shots of the dynamics are saved. 
\begin{figure}[t!]
    \centering
    \includegraphics[width=0.95\linewidth]{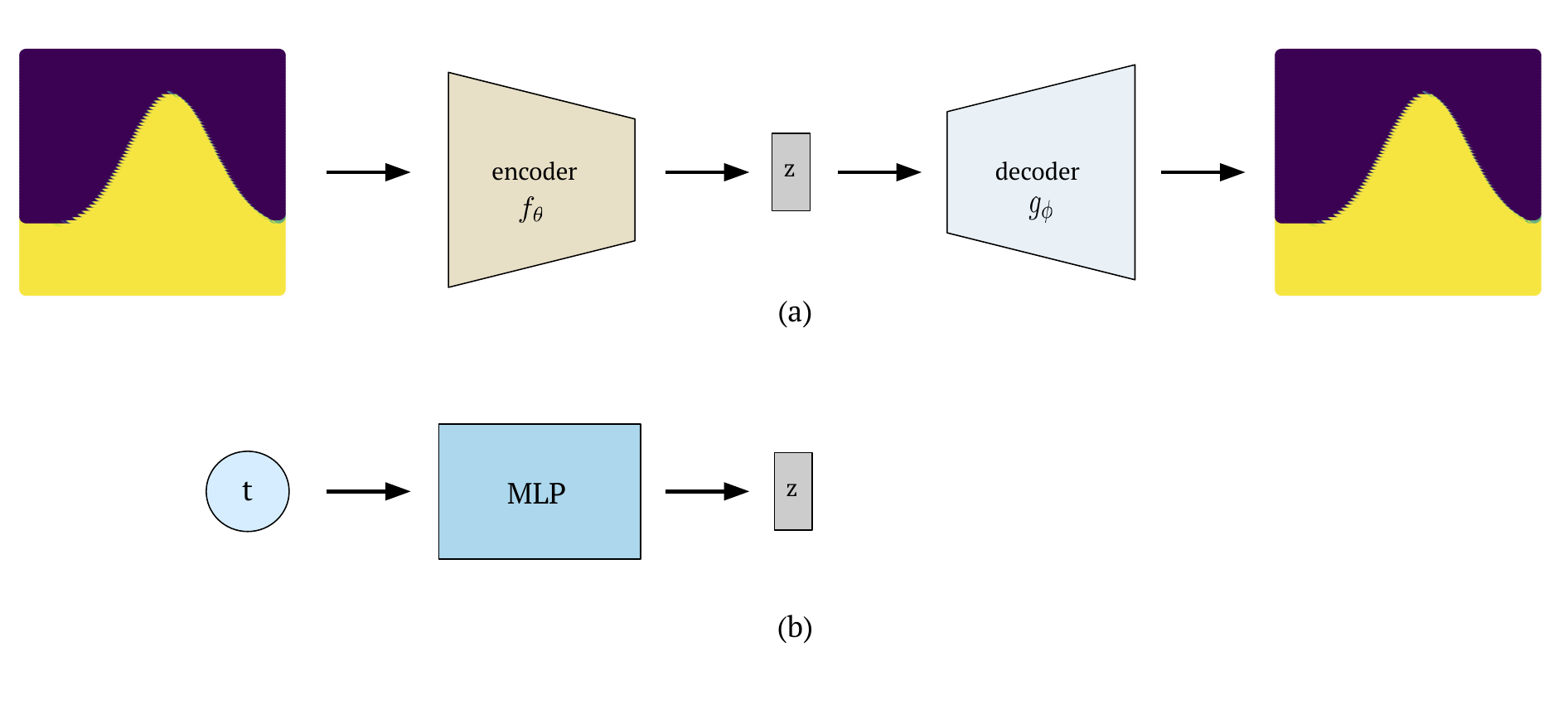}
    \caption{Multiphase network with $AE_\text{net}$ (a) and $TimeNet$ (b).}
    \label{multiphase_net}
\end{figure}
As depicted in Figure \ref{multiphase_net}, the overall deep learning architecture used (MP-Net) can be summarised into two building networks: an autoencoder ($AE_\text{net}$) to find the latent representation $z$; and a MLP ($TimeNet$) to map the time $t$ instance to the corresponding latent representation of the wave at time $t$. The $AE_\text{net}$ is composed by an encoder $f_\theta$ and a decoder  $g_\phi$. The encoder is a single continuous convolutional block with one input channel and output channel, stride and filter dimension both equal to $[0.4, 0.2]$, followed by a linear layer of dimension $(150, 30)$ with ELU \cite{elu} activation function. 
\begin{figure}[t!]
    \centering
    \includegraphics[width=0.95\linewidth]{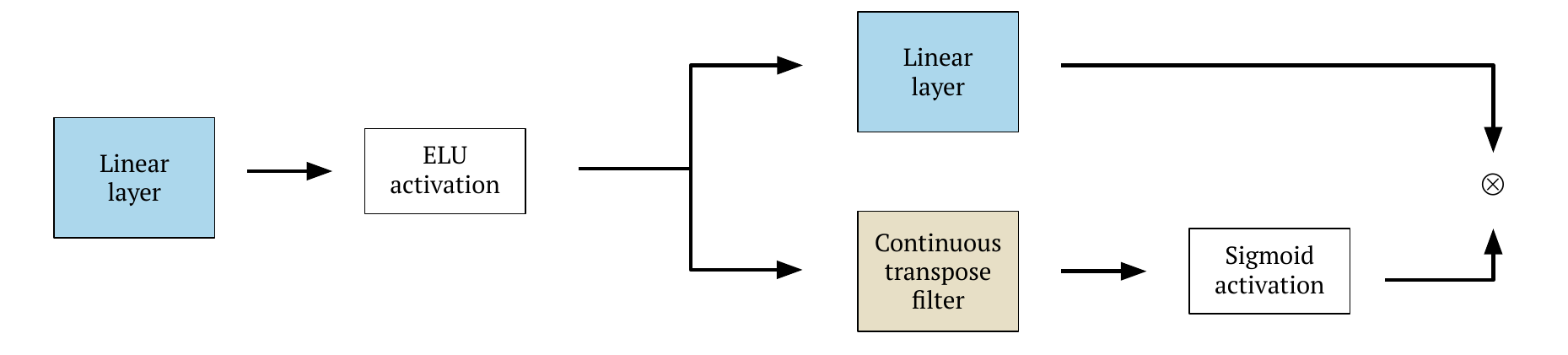}
    \caption{Decoder $g_\phi$ structure for the MP-Net.}
    \label{multiphase_decoder}
\end{figure}
The decoder, represented in Figure \ref{multiphase_decoder}, is composed by a linear layer of dimension $(30, 150)$, followed by ELU activation function. The output of the ELU layer is passed to two independent layers: firstly, a linear layer with dimension $(150, 18750)$; secondly, a transpose continuous filter layer of one input and output channel, stride and kernel size both equal to $0.4$, followed by a sigmoid activation function. The results of these two independent layers are multiplied, followed by an adaptive sigmoid activation function \cite{adaptive_functions}, which gives the output of the autoencoder. Both continuous filters, direct and transposed, are internally approximated by a MLP of dimension $10, 40, 80$ with an adaptive sigmoid activation function. The choice of multiplying the outputs of the two independent networks is due to the fact that a simple MLP decoder tends to over-fit, resulting in multiple waves in the output; while the continuous transpose filter tends to create a noisy output, resulting in a poor boundary approximation. By passing the continuous transpose filter representation to a sigmoid activation, we consider the output of the filter as a probability weight. Thus, each coordinate of the linear layer output is multiplied with this weight, with the intention of zeroing the noisy extra waves wrongly produced by a linear layer decoder, and keeping the wave of interest. Finally, the $TimeNet$ is a simple MLP of size $40, 80$ with a ReLU activation function. 

\begin{figure}[thb!]
    \centering
    \includegraphics[width = 0.95\linewidth]{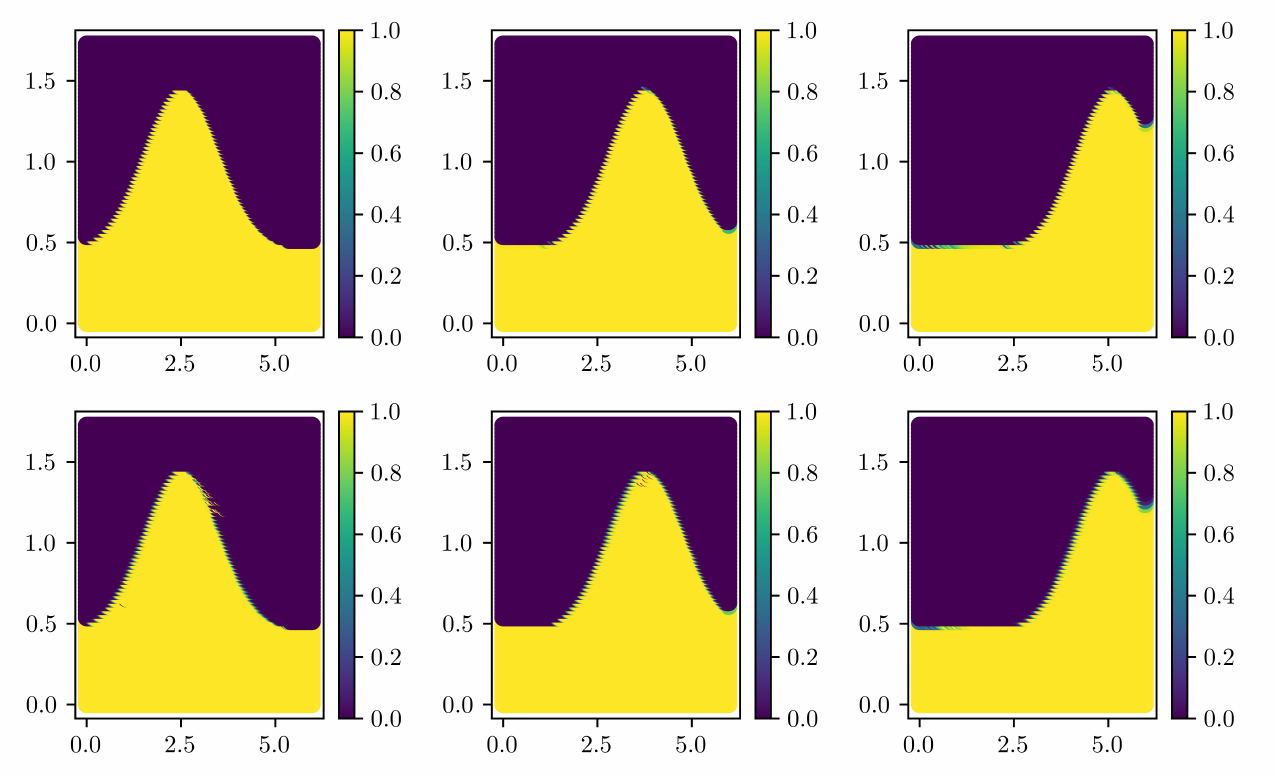}
    \caption{Samples of predicted snapshots (below) comparing to the originals (above) at different time steps}
    \label{inference_multiphase}
\end{figure}
\begin{figure}[thb!]
    \centering
    \includegraphics[width=0.95\linewidth]{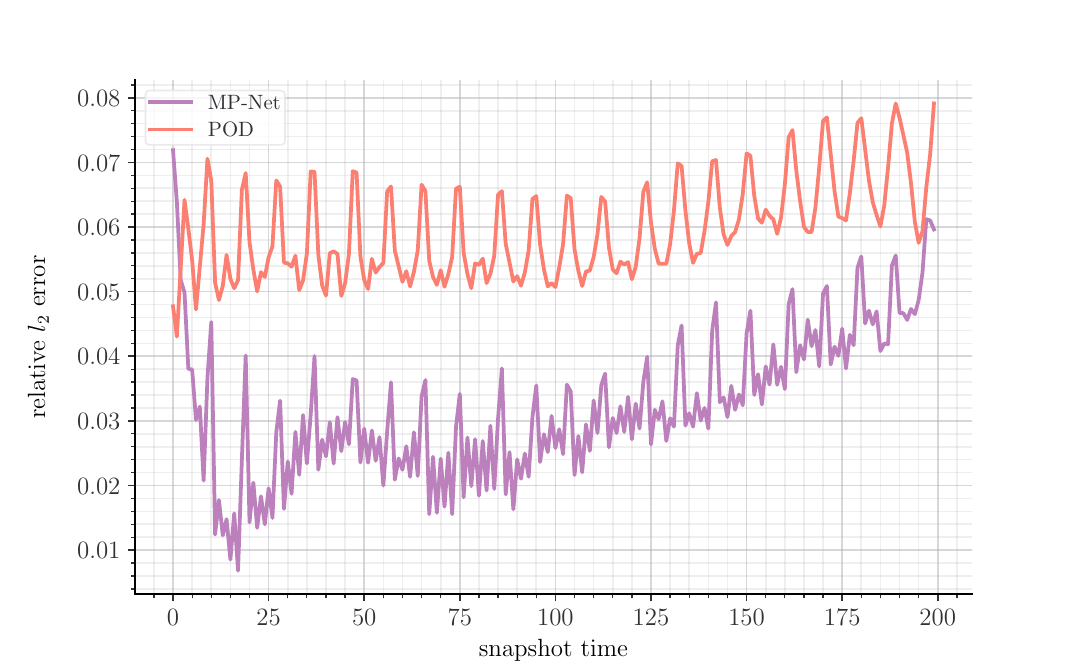}
    \caption{Variation of $l_2$ relative error for different snapshot for POD and MP-Net.}
    \label{error_l2_mpnet}
\end{figure}

Training is done on $90$ snapshots equally spaced in time, while testing is done on the remaining snapshots. The training process is divided in two parts as well: first training $AE_\text{net}$ for obtaining a latent representation $z$ of the input $I$ through encoding $z = f_\theta(I)$ and decoding $g_\phi(z)$; second, once $AE_\text{net}$ is completed, training $TimeNet$ to map a time step $t$ to its corresponding latent representation $z$ by using a MLP, minimizing the mean square error between $z$ and the $TimeNet$ output. $AE_\text{net}$ is trained using Adam optimizer \cite{adam} with initial learning rate $0.001$ and using a exponential learning rate scheduler with multiplicative factor of learning rate decay equal to $0.99$. The training is done on $300$ epochs with data distributed in batches of $5$, minimizing the mean square error loss. $TimeNet$ is trained using Adam optimizer with learning rate $0.001$, minimizing the mean square loss for $10000$ epochs with a batch size of $5$.

During inference, given only a time $t$ the snapshot is constructed. In particular, for each $t$ a latent representation $z$ is obtained using $TimeNet$, thus the former is passed to $g_\phi$ decoder for getting the snapshot prediction. Notice that at inference time the encoder is not used since the snapshot samples are not known during inference. Samples of predicted snapshots at different times are reported in Figure \ref{inference_multiphase}.
Overall, on the $110$ unseen snapshots the network has an $l_2$ relative error of $2.7$ on training, and $3.6$ on testing. The obtained test error is notable considering the fact that the network is equation-agnostic, and only $90$ samples are used for training. As a comparison baseline, the Proper Orthogonal Decomposition (POD), \reviewerA{a possible technique for model order reduction}, has been employed. By setting the same latent dimension size and snapshot used for MP-Net, and by using a radial basis function interpolator, the POD leads to a relative $l_2$ test error of $6.1$ percentage.

In Figure \ref{error_l2_mpnet}, we report the variation of $l_2$ relative error for all the snapshots (training + testing) for POD and MP-Net. The graph highlights that MP-Net outperforms POD for almost all snapshots, showing that MP-Net is able to find a better latent representation than a classical POD.

\section{Conclusion}\label{sec4}
We propose a continuous trainable filter to perform convolution on unstructured data. We demonstrate the validity of the methodology on different machine learning frameworks, such as supervised and self-supervised learning, and on different type of data e.g., images (structured) and meshes (unstructured). We show that a continuous filter is able to reproduce the same results as the state-of-the-art discrete filter on structured data, and it can be used efficiently in deep learning architectures with unstructured data to reproduce unseen snapshots in a fluid dynamics simulation. In addition, we show that by using continuous convolutional autoencoders higher performances can be achieved with respect to the more simpler feed-forward autoencoder, when unstructured data are used. Possible further investigation include finding the most performing inner network architecture in the continuous convolutional filter, or using the latter for transfer learning representations: indeed, by being independent on the input image size, incompatible data-sets can easily be merged. Furthermore, possible researches using the continuous filter for image inpainting algorithms could be done, considering the shown ability of the filter to work with partially completed images. Moreover, extending the framework to a 3-dimensional convolution might open the possibility to apply the continuous convolution to more complex manifolds. Finally, the use of the presented continuous filter in physics informed neural network architectures might be investigated, especially for problems where data are not structured. 
The architecture could be also applied to projection-based \cite{StabileZancanaroRozza2020,GeorgakaStabileRozzaBluck2019}  and non-intrusive reduced order models \cite{TezzeleDemoStabileMolaRozza2020} extending, for example, the work done in \cite{romor_ae} to unstructured meshes. 
To summarize, the presented methodology, combined with the proposed research directions, opens the possibility to use the continuous convolutional filter in very complex settings, spacing in several application fields of science and industry.

\section{Acknowledgements}
This work is partially supported by European Union Funding for Research and Innovation — Horizon 2020 Program — in the framework of European Research Council Executive Agency: H2020 ERC CoG 2015 AROMA-CFD project 681447 "Advanced Reduced Order Methods with Applications in Computational Fluid Dynamics" P.I. Professor Gianluigi Rozza, by European Union Funding for Research and Innovation --- Horizon Europe Program --- in the framework of European Research Council Executive Agency: ERC POC 2022 ARGOS project 101069319 ``Advanced Reduced order modellinG: Online computational web server for complex parametric Systems'' P.I. Professor Gianluigi Rozza, by European High-Performance Computing Joint Undertaking project Eflows4HPC GA N. 955558, by PRIN "Numerical Analysis for Full and Reduced Order Methods for Partial Differential Equations" (NA-FROM-PDEs) project.

\newpage
\bibliographystyle{abbrv}
\bibliography{biblio}

\end{document}